\documentclass[lettersize,journal]{IEEEtran}
\usepackage{amsmath,amsfonts}
\usepackage{algorithmic}
\usepackage{algorithm}
\usepackage{array}
\usepackage[caption=false,font=normalsize,labelfont=sf,textfont=sf]{subfig}
\usepackage{textcomp}
\usepackage{stfloats}
\usepackage{url}
\usepackage{verbatim}
\usepackage{graphicx}
\usepackage{cite}

\usepackage{algorithm}
\usepackage{algorithmic}
\usepackage{threeparttable}
\usepackage{bbding}
\usepackage{utfsym}
\usepackage{amsmath}
\usepackage{color}
\usepackage{xcolor}         
\usepackage{colortbl}
\definecolor{tabgray1}{gray}{.9}
\definecolor{tabgray2}{gray}{.84}
\usepackage{multirow} 
\usepackage{makecell}
\usepackage{pifont}
\definecolor{green}{rgb}{0.0,1.0,0.0}

\definecolor{redd}{rgb}{1.0,0.0,0.0}

\usepackage{bm}
\definecolor{mygray}{gray}{.7}
\definecolor{mypink}{rgb}{.99,.91,.95}
\definecolor{mycyan}{cmyk}{.3,0,0,0} 
\usepackage{calligra}
\usepackage{color}
\usepackage{amssymb}
\definecolor{BlueColor}{rgb}{0.0, 0.0, 1.0}
\definecolor{RedColor}{rgb}{1.0, 0.0, 0.0}
\definecolor{BlackColor}{rgb}{0.0, 0.0, 0.0}
\definecolor{GreenColor}{rgb}{0.3, 0.6, 0.4}

\newcommand{\hong}[1]{{\color{BlackColor} #1}}
\newcommand{\hongSwitch}{\color{BlackColor}}
\usepackage{lineno}
\usepackage{booktabs}
\usepackage{xspace} 
\newcommand{\eg}{\emph{e.g.}\xspace}
\newcommand{\ie}{\emph{i.e.}\xspace}
\newcommand{\etal}{\emph{et al.}\xspace}

\usepackage{newfloat}
\usepackage{listings}
\usepackage{bibentry}

\begin{document}
\title{$\text{M}^2\text{C-EvDet}$: Multi-Domain Multi-Order Cross-Modal Knowledge Distillation for Event-based Object Detection}

\author{

Wei Bao, Siqi Li, Shouan Pan, Yi Xie and Yue Gao, \IEEEmembership{Senior Member, IEEE}

\thanks{Wei Bao, Siqi Li, Shouan Pan and Yue Gao are with BNRist, THUIBCS, BLBCI, School of Software, Tsinghua University, Beijing 100084, China, and Yangtze Delta Region Institute, Tsinghua University, Jiaxing 314006, China (\{baoweivvv, lisiqi19971013, kevin.gaoy\}@gmail.com; psa24@mails.tsinghua.edu.cn).}

\thanks{Yi Xie is with School of Economics and Management, Beijing Forestry University, Beijing 100083, China (yixie@bjfu.edu.cn).}

\thanks{This work was supported in part by Brain Science and Brain-like Intelligence Technology—National Science and Technology Major Project (2025ZD0217300), National Natural Science Foundation of China (No. U25A20532), Project under the Key R\&D and Transformation Program of Qinghai Province (Competition-based Funding Model): Research and Demonstration on Early Warning and Prevention Technologies for Human-Bear Conflicts in Qinghai Province (Phase III). (\textit{Corresponding author: Siqi Li.})}

}

\maketitle
\begin{abstract}

Event-based object Detection (EvDet), as a biologically inspired visual perception paradigm, demonstrates superior performance in scenarios demanding high temporal resolution and a wide dynamic range. Nevertheless, the inherent sparse representations and inadequate visual semantics of event data result in a considerable performance disparity between EvDet and frame-based object detection. Previous works attempt to alleviate this cross-modal discrepancy through knowledge distillation, yet they only focus on spatial visual semantics or pair-wise relational information, thus limiting performance in more complex scenarios. To address this challenge, this paper proposes $\text{M}^2\text{C-EvDet}$, a Multi-domain and Multi-order Cross-modal knowledge distillation framework for EvDet. Built upon frequency learning and hypergraph computation, $\text{M}^2\text{C-EvDet}$ integrates two specialized modules: Adaptive Frequency-Decoupled Feature Distillation ($\text{AF}^2\text{D}^2$) and Multi-Order Relational Distillation ($\text{MORD}$). Specifically, $\text{AF}^2\text{D}^2$ conducts dynamic low- and high-frequency feature distillation via adaptive frequency decoupling, which eliminates the aliasing between modality-agnostic and modality-specific knowledge and thus enhances the transfer of global and local object semantics. $\text{MORD}$ transfers both pair-wise low-order relations and multi-to-multi high-order relations via self-attention and hyper-attention mechanisms, respectively, thereby further extending the capability boundaries of feature distillation. Concretely, the hyper-attention employs an aggregation-and-distribution paradigm, where hyperedge features are utilized as an intermediate bridge to enable high-order information propagation among multiple feature nodes, thus effectively capturing intricate high-order relational knowledge. Extensive experiments on three RGB-Event object detection datasets demonstrate the effectiveness of our method. Specifically, $\text{M}^2\text{C-EvDet}$ achieves 3.6 mAP improvement and establishes a new state-of-the-art benchmark on the DSEC-Detection dataset.
\end{abstract}

\section{Introduction}
\label{sec:intro}
Event cameras are a new type of bio-inspired vision sensor where each pixel asynchronously responds to brightness changes~\cite{survey}. Let $I(x,y,t)$ denote the brightness of pixel $(x,y)$ at timestamp $t$. When the logarithmic change of the brightness exceeds a specific threshold $C$, \ie, $|\log(I(x,y,t))-\log(I(x,y,t'))|>C$, an event will be triggered, where $t'$ is the timestamp of the previous event triggered at this pixel. The triggered event is denoted as $e=(x,y,t,p)$, where $t$ is the timestamp and $p\in\{1,-1\}$ is the polarity, indicating whether the brightness is increased or decreased. All events triggered asynchronously by all pixels of the event camera converge to form the output event stream. Due to their unique working principle, event cameras have advantages such as high temporal resolution (about 1 $\mu$s), high dynamic range (about 140 dB), and low latency, making them widely used in fields like sign language recognition~\cite{jiang2024evcslr}, autonomous driving~\cite{li2025rgb,jiang2023event,e3dt}, feature tracking~\cite{e3dt}.

The high temporal resolution and high dynamic range characteristics have driven widespread attention to \textbf{Ev}ent-based Object \textbf{Det}ection (EvDet) in the computer vision community. Over the past few years, numerous state-of-the-art event feature extraction networks, ranging from GNN-based models~\cite{schaefer2022aegnn} and SNN-based models~\cite{zhang2022spiking} to CNN/Transformer-based architectures~\cite{gehrig2023recurrent,torbunov2025evrt}, have been developed to facilitate object perception tasks in complex scenes. RVT~\cite{gehrig2023recurrent} proposes local-global self-attention and recurrent temporal attention to aggregate spatial and temporal information, respectively. EvRT-DETR~\cite{torbunov2025evrt} transforms image-based detectors into event-based detection models by modifying their frozen latent representation space in an efficient adaptation way. However, constrained by the intrinsic sparsity of event data and the lack of visual semantic information, EvDet approaches suffer from a significant performance discrepancy compared with conventional RGB frame-based detection methods. To address this challenge, DAGr~\cite{gehrig2024low} leverages the high temporal resolution and inherent sparsity of event data and the rich but low temporal resolution information of RGB images to achieve efficient and high-performance object detection. Subsequently, a multitude of RGB-Event fusion approaches~\cite{lu2412flexevent,zhang2024frequency} have been developed to fully leverage the complementary properties of multi-modal data. However, these methods still necessitate strictly synchronized RGB-event data pairs during the inference phase, which significantly restricts their applicability to scenarios where RGB cameras are unavailable.
\begin{figure*}
\begin{center}
\includegraphics[width=1.0\linewidth]{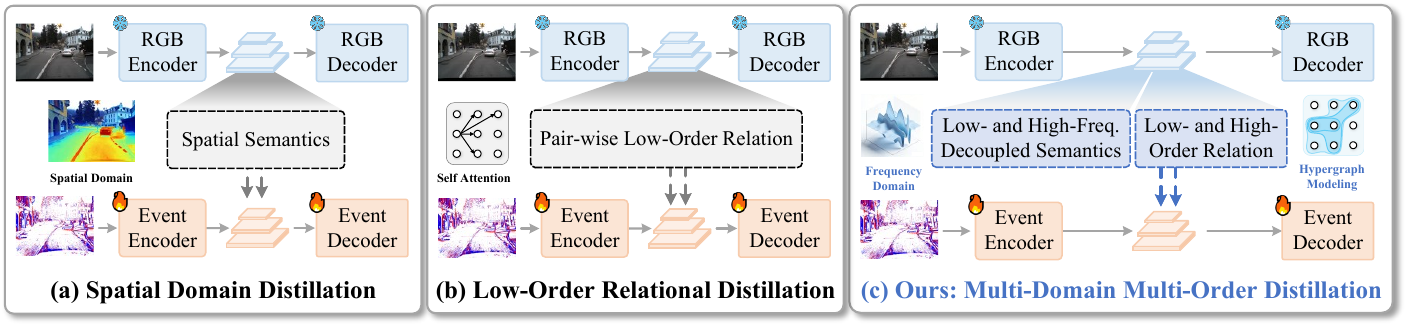}
\end{center}
\vspace{-0.5cm}
\caption{\textbf{Comparison of our proposed $\text{M}^2\text{C-EvDet}$ with existing methods}, including (a) whole spatial domain distillation and (b) low-order relational distill. Our method (c) transfers low-frequency and high-frequency decoupled object semantics, and low-order and high-order object relations through frequency learning and hypergraph computation.}
\vspace{-0.5cm}
\label{fig1:intro}
\end{figure*}

\hong{To eliminate the dependency on RGB data during the inference stage, recent studies have embraced cross-modal knowledge distillation techniques, which enable the exploitation of the rich visual semantics inherent in RGB data~\cite{zhang2023global,zhang2023channel,zhang2025closer,zhang2025reliable} during the training phase.} COFD~\cite{li2024object} proposes an object-centric slot attention mechanism that enables effective object-centric spatial distillation and pairwise relational distillation. EA-DETR~\cite{rossi2025event} leverages the sparsity of event data to generate an event-aware binary mask that identifies object-relevant regions and facilitates the transfer of more useful spatial semantics. It can be observed that existing cross-modal knowledge distillation methods for EvDet still perform object-centric knowledge distillation within a single spatial domain shown in Fig.~\ref{fig1:intro}(a) or simple pair-wise low-order relational distillation shown in Fig.~\ref{fig1:intro}(b), leading to insufficient information transfer capability in complex scenarios. Consequently, a more holistic and robust framework is critically required to facilitate advanced visual knowledge transfer, overcoming the constraints of conventional distillation methods.

Based on the aforementioned analysis, this paper proposes $\text{M}^2\text{C-EvDet}$, a \textbf{M}ulti-domain \textbf{M}ulti-order \textbf{C}ross-modal knowledge distillation framework used for event-based Object \textbf{Det}ection as shown in Fig.~\ref{fig1:intro}(c). $\text{M}^2\text{C-EvDet}$ leverages frequency learning~\cite{li2020wavelet} and hypergraph computation~\cite{hgnn} to enhance traditional spatial-domain feature distillation and pair-wise low-order relational distillation, leading to two key modules: \textbf{A}daptive \textbf{F}requency-\textbf{D}ecoupled \textbf{F}eature \textbf{D}istillation ($\text{AF}^2\text{D}^2$) and \textbf{M}ulti-\textbf{O}rder \textbf{R}elational \textbf{D}istillation ($\text{MORD}$). On the one hand, single spatial distillation methods tend to induce aliasing between modality-agnostic and modality-specific knowledge when processing cross-modal data, leading to constrained object semantic transfer capability. To mitigate this issue, $\text{AF}^2\text{D}^2$ leverages frequency-domain knowledge to perform low- and high-frequency decoupled spatial distillation, which enhances the transfer of global and local object semantics, respectively. Furthermore, to accommodate objects with diverse appearances and sizes across images, we realize adaptive low- and high-frequency separation through multi-scale wavelet decomposition and adaptive wavelet fusion. On the other hand, conventional relational distillation approaches commonly utilize self-attention mechanisms to transfer object relational knowledge from teacher models. However, this pairwise low-order modeling framework struggles to cope with the sophisticated inter-object correlations present in complex real-world environments. To tackle this limitation, the proposed $\text{MORD}$ module, building upon low-order relational distillation, designs hyper-attention via hypergraph computation theory to model and transfer multi-to-multi high-order relations, further extending the capability boundaries of feature distillation. Specifically, the hyper-attention employs an aggregation-and-distribution paradigm, where hyperedge features are utilized as an intermediate bridge to enable high-order information propagation among multiple feature nodes. By adapting and integrating the above two components into object detection framework, $\text{M}^2\text{C-EvDet}$ can fully leverage the rich visual and texture information inherent in the RGB modality, thereby effectively enhancing the detection performance on event stream. The experimental results demonstrate the superior detection capabilities and state-of-the-art results of our proposed method. The main contributions of our paper are summarized as follows:
\begin{itemize}
    \item We propose $\text{M}^2\text{C-EvDet}$, a multi-domain multi-order cross-modal distillation framework for event-based object detection, fully harnessing the rich visual and texture information inherent in the RGB modality.
    \item An adaptive frequency-decoupled feature distillation is proposed to mitigate the aliasing between modality-agnostic and modality-specific knowledge through adaptive low- and high-frequency decoupled distillation,  boosting the transfer of global and local object semantics.
    \item A multi-order relational distillation is proposed to model and transfer both pair-wise low-order relations and multi-to-multi high-order relations through self-attention and hypergraph attention, further expanding the capability boundaries of feature distillation.
    \item We conduct comprehensive experiments on three RGB-guided event-based object detection datasets, which demonstrate that $\text{M}^2\text{C-EvDet}$ achieves superior performance and establishes state-of-the-art (SOTA) results.
\end{itemize}

\begin{figure*}
\begin{center}
\includegraphics[width=1\linewidth]{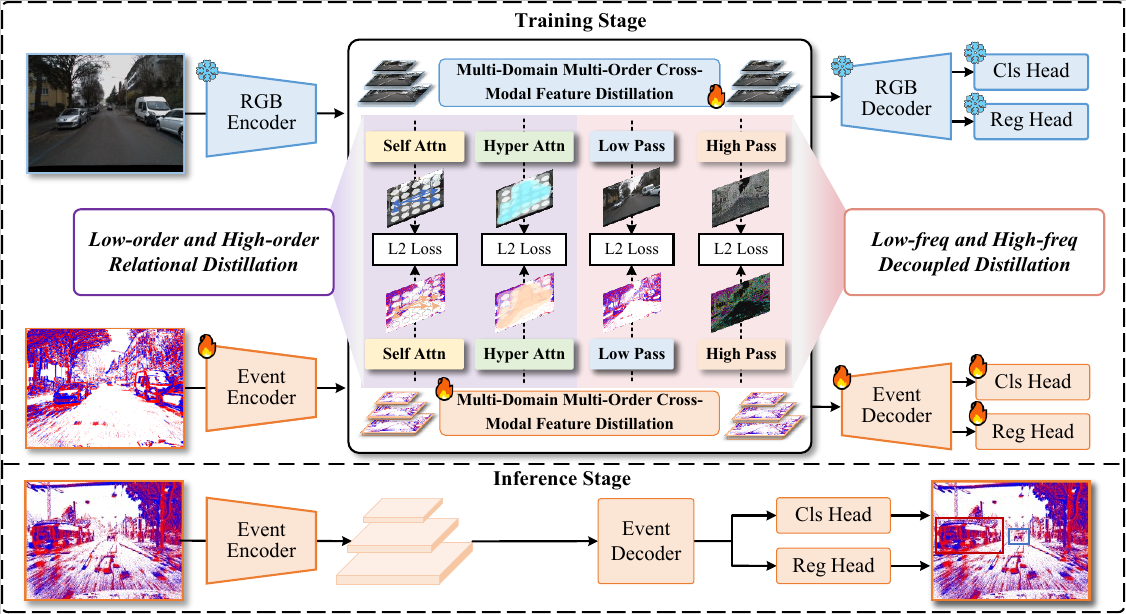}
\end{center}
\caption{
\textbf{Overview of the proposed $\text{M}^2\text{C}\text{-EvDet}$ framework.} During training phase, our method is built on a teacher-student architecture and takes event streams and RGB frames as inputs. The $\text{M}^2\text{CFD}$ module is employed to implement low-order and high-order relational distillation, and low-frequency and high-frequency decoupled distillation. During inference phase, our method solely take event streams as inputs to perform object detection.}
\label{fig2:overview}
\end{figure*}

\section{Related work}
\subsection{Event-Based Object Detection}
Event-based object detection methods can be broadly split into two categories, \ie, GNN/SNN-based models~\cite{schaefer2022aegnn,zhang2022spiking} and CNN/Transformer-based models~\cite{gehrig2023recurrent,torbunov2025evrt}. GNN/SNN-based methods directly take raw event stream as input and extract the visual feature from sparse event stream. Due to the advanced performance, CNN/Transformer-based frameworks have emerged as the dominant paradigm. These methods~\cite{gehrig2023recurrent,torbunov2025evrt} extract spatial- and temporal-wise features to enhance the event representation. However, they lack semantic richness and lead to a substantial performance gap between EvOD and frame-based methods. Recently, many event-frame fused methods~\cite{gehrig2024low,lu2412flexevent,zhang2024frequency} have attempted to use the high temporal resolution and inherent sparsity of event data, and the rich but low temporal resolution information of RGB images to deal with highly dynamic scenarios. To further break the limit of strictly synchronized RGB-event data pairs during the inference phase, COFD~\cite{li2024object} and EA-DETR~\cite{rossi2025event} leverage spatial or relational knowledge distillation which can transfer the rich object-relevant semantics and relations from RGB modality. A number of alternative RGB-guided event-based perception approaches~\cite{burkhardt2025superevent,bartolomei2025depth,liu2025i2ekd} also accomplish the transfer of rich visual and texture information from the RGB modality during the training phase.

Despite these advances, existing methods only focus on the knowledge in the single spatial domain or pair-wise low-order relations, leading to insufficient information transfer capability in complex scenarios. Differently, this work transfers low-frequency and high-frequency decoupled object semantics, and low-order and high-order object relations based on frequency learning and hypergraph computation.
\subsection{Knowledge Distillation for Object Detection}
Knowledge distillation~\cite{li2023object,yang2023context,fu2025hierarchical} transfers rich knowledge from large-scale detectors to compact student detectors for model compression and acceleration. FKD~\cite{zhang2020improve} proposes to learn not only the feature of an individual pixel but also the relation between different pixels captured by non-local modules. FGD~\cite{yang2022focal} decouples features into focal and global distillation, capturing fine-grained object features and global context to mitigate background interference. MGD~\cite{yang2022masked} makes the student model generate the teacher feature with its masked feature instead of mimicking it directly. Additionally, several distillation methods~\cite{chen2022d,chang2023detrdistill,wang2024kd,lan2025clockdistill}, tailored for DETR models leverage query-related information to enhance the performance of feature distillation or logit distillation. DETRDistill~\cite{chen2022d} proposes a target-aware feature distillation strategy to enable the student model to learn from the object-centric features extracted by the teacher model. CLoCKDistill~\cite{lan2025clockdistill} distills the transformer encoder output, which encapsulates valuable global context and long-range dependencies. Notably, our method can leverage adaptive frequecny decoupling and hyper-attention to model and transfer global and local semantics, and high-order relational knowledge.
\subsection{Frequency learning in Computer Vision}

Images in the spatial domain can be transformed into frequencies in the spectral domain; conversely, frequencies can be losslessly converted back to images. Frequency domain analysis~\cite{li2020wavelet,chen2024frequency} has been widely applied in various computer vision tasks. WD-DETR~\cite{cui2025wd} integrates wavelet transform into the backbone network to filter noise, as event representations or images can be easily treated as signals. In multi-modal images, low-frequency features primarily contain modality-agnostic information that is highly shared across different modalities, while high-frequency features tend to capture modality-specific information and are usually accompanied by more noise. FAD~\cite{pham2024frequency} employs a global filter in the frequency domain to adjust the frequencies of the student’s features under the guidance of the teacher’s features, failing to fully utilize frequency components for cross-modal knowledge transfer. FreeKD~\cite{zhang2024freekd} generates pixel-wise frequency masks via Frequency Prompt to localize pixels of interest across various frequency bands. However, the frequency prompt is learned through the teacher network, which is prone to overfitting in cross-modal images. Unlike these methods, our approach fully leverages frequency-domain information to decouple the aliasing of low and high frequencies in cross-modal knowledge transfer, enhancing the global and local semantic information.

\subsection{Visual Hypergraph Computation}
Hypergraph is the extension of graph. A hypergraph $\mathcal G$ can be defined by the hyperedge set $\mathcal E$ and the vertex set $\mathcal V$, \ie, $\mathcal G=\{\mathcal E, \mathcal V\}$. Different from traditional graphs, each hyperedge of a hypergraph can connect multiple vertices, and thus complex correlations between multiple vertices can be modeled. Hypergraph learning can facilitate the correlation-guided analysis by capturing complex and higher-order correlations in data based on hypergraph structures. Gao~\etal~\cite{hgnn} propose a higher-order message spatial propagation method between vertices, which further enhance the capability of hypergraph learning. Recently, several works~\cite{ma2021learning,11275932,xu2025hypergraph,xu2025residual} have applied hypergraph computation to computer vision. HyperYOLO~\cite{feng2024hyper} is the first work to apply hypergraph computation to object detection, proposing a semantic collecting and scattering framework and modeling hyperedges as $\epsilon$-ball centered at each feature point. YOLOv13~\cite{lei2025yolov13} advances hyperedge modeling by formulating them as global features, where each feature is generated via learnable prototypes combined with global offsets to facilitate hypergraph message propagation in a soft-connection manner. Notably, our method is the first work to apply hypergraph computation to knowledge distillation for object detection, designing hyper attention that acquires hyperedge features via K-means clustering to further transfer high-order relational knowledge.

\section{Method}
\label{sec:method}
 In this section, we introduce our proposed $\text{M}^2\text{C}\text{-EvDet}$, an efficient RGB-guided cross-modal distillation framework for EvDet. The overview of our proposed method is first introduced in Sec.~\ref{sec:3.1}. Then, Sec.~\ref{sec:3.2} and Sec.~\ref{sec:3.3} introduces the $\text{AF}^2\text{D}^2$ module and $\text{MORD}$ module, respectively. Finally, we give the overall loss in Sec.~\ref{sec:3.4}.
 
\subsection{Overview}
\label{sec:3.1}
Fig.~\ref{fig2:overview} depicts the overall pipeline of our proposed $\text{M}^2\text{C}\text{-EvDet}$. During the training phase, $\text{M}^2\text{C}\text{-EvDet}$ takes event streams and RGB frames as inputs, adopting a teacher-student architecture to transfer knowledge from RGB frames to event data. Then, the multi-scale RGB frame features and event features are extracted by the RGB encoder and event encoder (typically composed of a backbone network plus FPN~\cite{lin2017feature}/PAFPN~\cite{liu2018path}), respectively. Subsequently, our method employs Multi-Domain Multi-Order Cross-Modal Feature Distillation ($\text{M}^2\text{CFD}$), which consists of $\text{AF}^2\text{D}^2$ and $\text{MORD}$, to distill the rich visual and texture information inherent in the RGB modality. Specifically, $\text{MORD}$ adopts self-attention and hyper-attention to implement low-order and high-order relational distillation, respectively. $\text{AF}^2\text{D}^2$ employs low-pass and high-pass filtering modules to perform decoupled distillation of low-frequency and high-frequency components. During the inference stage, our method solely takes event streams as inputs and leverages the enhanced the multi-scale event features to achieve improved detection performance. It is worth noting that our method is a feature distillation approach, which can be applied to various detection frameworks. 
\subsection{Adaptive Frequency-Decoupled Feature Distillation}
\label{sec:3.2}
$\text{AF}^2\text{D}^2$ is designed to decouple conventional whole spatial-domain feature distillation, overcoming its inherent performance limitations by leveraging frequency-domain analysis. For multimodal images, low-frequency features primarily encode modality-agnostic information, which is highly shared across different modalities. In contrast, high-frequency features tend to capture modality-specific information and are typically accompanied by more noise. Directly distilling semantic information from the spatial domain tends to result in the conflation of modality-general and modality-specific knowledge. Therefore, $\text{AF}^2\text{D}^2$ leverages frequency-domain features to conduct separate low-frequency and high-frequency feature distillation. Low-frequency distillation excludes modality-induced noise and learns global modality-shared knowledge, whereas high-frequency distillation acquires sharp object contours from RGB images and suppresses modality-specific noise components. In addition, objects exhibit variations in size and morphology across different images, and a single-scale high-low frequency decomposition strategy struggles to efficiently decouple the global and local semantics of targets. Consequently, we design a adaptive frequency-domain decomposition mechanism, enabling more effective high-low frequency distillation via multi-scale wavelet decomposition and adaptive wavelet fusion. 
\begin{figure}[!t]
\begin{center}
\includegraphics[width=0.9\linewidth]{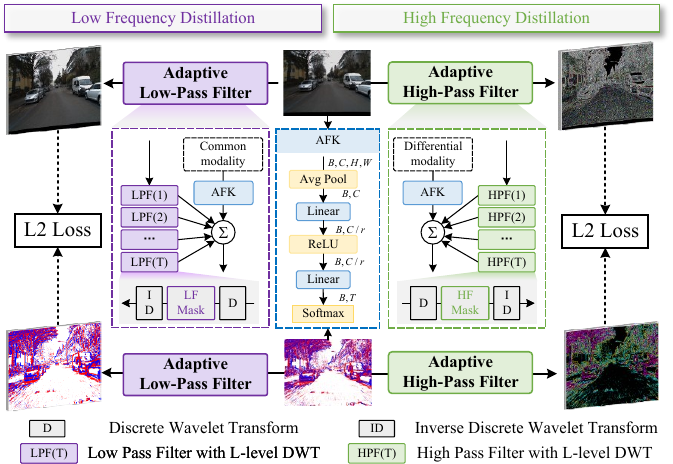}
\end{center}
\vspace{-0.3cm}
\caption{\textbf{Illustration of the $\text{AF}^2\text{D}^2$ module}, which leverages adaptive low-pass and high-pass filter to achieve the decoupling of low-frequency and high-frequency feature distillation.}
\vspace{-0.5cm}
\label{fig3:freq}
\end{figure}

As illustrated in Fig.~\ref{fig3:freq}, adaptive low-pass filtering and adaptive high-pass filtering are adopted to realize low-frequency and high-frequency decomposition. The adaptive low-pass filtering component incorporates multi-level Low-Pass Filters (LPFs), while the adaptive high-pass filtering component is composed of multi-level High-Pass Filters (HPFs). To instantiate these filtering operations, we utilize the Discrete Wavelet Transform (DWT), low-/high-frequency masking mechanisms, and Inverse Discrete Wavelet Transform (IDWT) to achieve the LPF and HPF functionalities. Supposing that the multi-scale features of RGB modality after PAFPN is denoted as $\mathcal{S}_l \in \mathbb{R}^{C_l \times H_l \times W_l}$ and that of event modality is denoted as $\mathcal{T}_l \in \mathbb{R}^{C_l \times H_l \times W_l}$, where $l=3,4,5$. $C_l$, $H_l$, and $W_l$ is the channel dimension, height and width of $l$-th feature maps, respectively. For each level, we first adopt the spatial-to-frequency domain transformation to obtain frequency components, \ie,
\begin{equation}
    \hat{\mathcal{S}_l^t} = \text{DWT}^t(\mathcal{S}_l), \quad \hat{\mathcal{T}_l^t} = \text{DWT}^t(\mathcal{T}_l),
\end{equation}
where $\hat{\mathcal{S}_l^t}$ and $\hat{\mathcal{T}_l^t}$ denote the transformed frequency-domain values after the $t$-th level DWT. When the decomposition level $t$ is set to 1, the feature map can be partitioned into four frequency bands ${\text{LL}^1,\text{HL}^1,\text{LH}^1,\text{HH}^1}$, where $\text{LL}^1$ denotes the low-frequency band and the remaining three correspond to high-frequency bands. When the decomposition level $t$ is set to 2, the $\text{LL}^1$ band can be further decomposed into four sub-bands ${\text{LL}^2,\text{HL}^2,\text{LH}^2,\text{HH}^2}$. We define the low-frequency mask operation $\mathcal{M}_t^{lf}$ as setting all high-frequency bands to 0 for the \textit{t}-th level DWT, and the high-frequency mask operation $\mathcal{M}_t^{hf}$ as setting the low-frequency band to 0 for the \textit{t}-th level DWT. Hence, the high-frequency and low-frequency components of the RGB teacher features and Event student features in the $t$-th level can be formulated as:
\begin{equation}
\begin{aligned}
\hat{\mathcal{T}}_l^{hf(t)} &= \mathcal{M}_{hf}^{t}( \hat{\mathcal{T}}_l^t), \quad \hat{\mathcal{T}}_l^{lf(t)} = \mathcal{M}_{lf}^{t}( \hat{\mathcal{T}}_l^t) \\
\hat{\mathcal{S}}_l^{hf(t)} &= \mathcal{M}_{hf}^{t}( \hat{\mathcal{S}}_l^t), \quad \hat{\mathcal{S}}_l^{lf(t)} = \mathcal{M}_{lf}^{t}( \hat{\mathcal{S}}_l^t) 
\end{aligned}.
\end{equation}
Finally, we use frequency-to-spatial domain transformation to obtain the reconstructed low-frequency and high-frequency features, \ie,
\begin{equation}
\begin{aligned}
\mathcal{T}_l^{hf(t)} &= \text{IDWT}^t(\hat{\mathcal{T}}_l^{hf(t)}), \quad \mathcal{T}_l^{lf(t)} = \text{IDWT}^t(\hat{\mathcal{T}}_l^{lf(t)}) \\
\mathcal{S}_l^{hf(t)} &= \text{IDWT}^t(\hat{\mathcal{S}}_l^{hf(t)}), \quad \mathcal{S}_l^{lf(t)} = \text{IDWT}^t(\hat{\mathcal{S}}_l^{lf(t)})
\end{aligned}.
\end{equation}
Subsequent to multi-scale wavelet decomposition, this work proposes an adaptive wavelet fusion module that leverages Attention over Filter Kernels (AFK) to aggregate multi-scale wavelet components in accordance with their respective attention weights. In multimodal image data, low-frequency components encapsulate modality-general knowledge that encodes the common characteristics across diverse modalities, while high-frequency components embody modality-specific knowledge that captures the discriminative traits unique to individual modalities. Building upon the shared properties and discriminative discrepancies inherent in multimodal images, AFK adopts a two-layer linear network to output attention scores for low-frequency features $\mathcal{A}_l^{lf}$ and attention scores for high-frequency features $\mathcal{A}_l^{hf}$, as follows:
\begin{equation}
\begin{aligned}
\mathcal{A}_l^{lf} &= \text{AFK}(\mathcal{T}_l + \mathcal{S}_l) \\
\mathcal{A}_l^{hf} &= \text{AFK}(\mathcal{T}_l - \mathcal{S}_l) \\
\text{AFK}(F) &= \text{SM}\left( \text{ReLU}(\text{AvgPool}(F)W_1)W_2\right)
\end{aligned},
\end{equation}
where $\text{AvgPool}$ is the average pooling, $\text{ReLU}$ is the relu activation function and $\text{SM}$ is the softmax activation function. $W_1 \in \mathbb{R}^{C_l \times C_l/r}$ and  $W_2 \in \mathbb{R}^{C_l/r\times T}$ are weights of linear layers, where $T$ is the number of DWT levels and $r$ is reduction ratio. We use $\mathcal{T}_l + \mathcal{S}_l$ to represent common modality and $\mathcal{T}_l - \mathcal{S}_l$ to represent differential modality. The final high-frequency and low-frequency components of the RGB teacher features and Event student features can be obtained as:
\begin{equation}
\begin{aligned}
\mathcal{S}_l^{lf} &= \sum_{t=1}^{T}\mathcal{A}_l^{lf(t)} \cdot \mathcal{S}_{l}^{lf(t)}, \quad
\mathcal{S}_l^{hf} = \sum_{t=1}^{T}\mathcal{A}_l^{hf(t)} \cdot \mathcal{S}_{l}^{hf(t)}\\
\mathcal{T}_l^{lf} &= \sum_{t=1}^{T}\mathcal{A}_l^{lf(t)} \cdot \mathcal{T}_{l}^{lf(t)}, \quad
\mathcal{T}_l^{hf} = \sum_{t=1}^{T}\mathcal{A}_l^{hf(t)} \cdot \mathcal{T}_{l}^{hf(t)}
\end{aligned}. 
\end{equation}
Finally, the distillation loss function $\mathcal{L}_{\text{freq}}$ in the frequency domain is obtained as follows:
\begin{equation}
\begin{aligned}
\mathcal{L}_{\text{freq}} &= \lambda_1 \cdot \mathcal{L}_{\text{freq}}^{lf} + \lambda_2 \cdot \mathcal{L}_{\text{freq}}^{hf} \\
\mathcal{L}_{\text{freq}}^{lf} &= \sum_{l \in \mathcal{L}} \sum \left(\mathcal{S}_{l}^{lf} - \mathcal{T}_{l}^{lf}\right)^2 \\
\mathcal{L}_{\text{freq}}^{hf} &= \sum_{l \in \mathcal{L}} \sum \left( \mathcal{S}_{l}^{hf} / \left\| \mathcal{S}_{l}^{hf} \right\|_2 - \mathcal{T}_{l}^{hf} / \left\| \mathcal{T}_{l}^{hf} \right\|_2 \right)^2
\end{aligned},
\end{equation}
where $\mathcal{L}$ is the number of feature maps levels, $\lambda_1$ and $\lambda_2$ are weight adjustment coefficients, $\mathcal{L}_{\text{freq}}^{lf}$ and $\mathcal{L}_{\text{freq}}^{hf}$ denote the high-frequency and low-frequency feature distillation losses, respectively. \( \|\cdot\|_2 \) denotes the \( L_2 \) norm. Since high-frequency features inherently contain substantial noise and are prone to generating abnormally large values, leading to excessive gradient magnitudes during training, normalization is exclusively applied to the high-frequency features.

\begin{figure}[t]
\begin{center}
\includegraphics[width=0.9\linewidth]{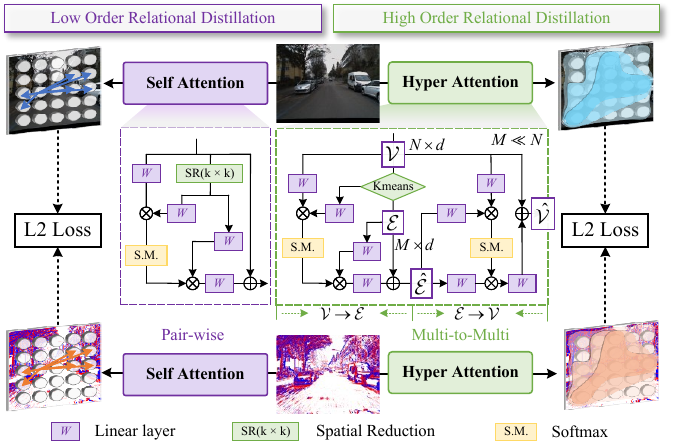}
\end{center}
\vspace{-0.3cm}
\caption{\textbf{Illustration of the MORD module}, which leverages self attention and hyper attention to transfer low-order and high-order relations.}
\vspace{-0.5cm}
\label{fig4:spatial}
\end{figure}
\subsection{Multi-Order Relation Distillation in Spatial Domain}
\label{sec:3.3}
The relational information between distinct pixels encapsulates valuable knowledge and has been leveraged to enhance the performance of detection tasks. Relation distillation has been proven effective in improving the performance of knowledge distillation by capturing the global correlational relationships of features~\cite{zhang2020improve}. Nevertheless, these approaches generally utilize non-local or attention-based modules to model pairwise associations between feature points, overlooking global multi-to-multi high-order correlations, which restricts distillation performance in complex scenarios. Our proposed $\text{MORD}$ leverages spatial-reduction self-attention and hypergraph attention based on hypergraph computation to separately model low-order and high-order relations among pixels, enabling the transfer of complex correlational structures from the teacher model to the student model. The low-order and high-order relations are retrieved via the following procedures, respectively. For the convenience of formula presentation in the relation distillation section, we reshape the tensor $\mathcal{S}_l \in \mathbb{R}^{C_l \times H_l \times W_l}$ and $\mathcal{T}_l \in \mathbb{R}^{C_l \times H_l \times W_l}$into $\mathcal{S}_l \in \mathbb{R}^{N_l \times C_l}$ and $\mathcal{T}_l \in \mathbb{R}^{N_l \times C_l}$, respectively. Here, $N_l=H_lW_l$ represents the number of all feature points in the $l$-th level feature map, and no further elaboration will be provided in subsequent content.

\textbf{Low-Order Relational Distillation.} Self-attention mechanisms have been extensively validated for their efficacy in modeling global associative relationships, yet they incur prohibitive computational costs in the context of high-resolution object detection scenarios. Several existing works have shown that partial spatial attenuation can effectively reduce computational complexity while still enabling efficient modeling of pairwise relationships. PVT~\cite{wang2021pyramid} achieves maximal retention of spatial information through spatial attenuation applied to the Key and Value tensors in the self-attention module.  Therefore, as illustrated in the Fig.~\ref{fig4:spatial}, this work adopts the spatial attenuation to model the low-order relationships from the teacher network. The loss function $\mathcal{L}_{\text{relation}}^{lr}$ of Low-Order Relational Distillation (LORD) is expressed as follows:
\begin{equation}
\begin{aligned}
\mathcal{L}_{\text{relation}}^{lr} &= \sum_{l \in \mathcal{L}} \sum \left(\mathcal{R}^{lr}(\mathcal{S}_l) - \mathcal{R}^{lr}(\mathcal{T}_l)\right)^2 \\
\mathcal{R}^{lr}(\mathcal{X}_l) &= \mathcal{X}_l+\left[\text{SM}\left(\frac{\mathcal{X}_lW_q^{lr} \ (\mathcal{X}_l^{sr}W_k^{lr})^{\top}}{\sqrt{C_l}}\right) \mathcal{X}_l^{sr}W_v^{lr}\right]W_o^{lr} \\
\mathcal{X}_l^{sr} &= \text{Re}^{C_lH_lW_l \to N_lC_l}(\text{Conv}_{k \times k}(\text{Re}^{N_lC_l\to C_lH_lW_l}(\mathcal{X}_l)))
\end{aligned},
\end{equation}
where $W_q^{lr} \in\mathbb{R}^{C_l \times C_l}$, $W_k^{lr}\in\mathbb{R}^{C_l \times C_l}$, $W_v^{lr}\in\mathbb{R}^{C_l \times C_l}$ and $W_o^{lr}\in\mathbb{R}^{C_l \times C_l}$ denote linear layers, $\mathcal{R}^{lr}(\cdot)$ captures the pair-wise low-order correlations, and $\sqrt{C_l}$ is to avoid gradient explosion. $\mathcal{X}_l^{sr}$ is the feature after spatial reduction which can be achieved by a convolution layers with the kernel size and stride both as $k \times k$. $\text{Re}^{C_lH_lW_l \to N_lC_l}$ denotes the reshape operation which convert the size of tensor from $C_l \times H_l \times W_l$ to $N_l \times C_l$. $\text{Re}^{N_lC_l \to C_lH_lW_l}$ is inverse operation. In addition, we incorporate a multi-head attention mechanism~\cite{vaswani2017attention} to augment the diversity of the captured relational correlations. 

\textbf{High-Order Relational Distillation.} LORD is limited to local information aggregation and pairwise correlation modeling, lacking the capability to capture global multi-to-multi high-order correlations. Hence, High-Order Relational Distillation (HORD) is proposed to model high-order correlations via hypergraph computation. Hypergraph is the extension of graph, and can be defined by the hyperedge set $\mathcal E$ and the vertex set $\mathcal V$, \ie, $\mathcal G=\{\mathcal E, \mathcal V\}$.  In contrast to conventional graph edges that are restricted to linking two vertices, a hyperedge can connect multiple vertices in a single association. Thus, hypergraph could model high-order correlations between multiple vertices. HORD designs the hyper-attention which adopts an aggregation-and-distribution paradigm, leveraging adaptive hyperedge features as an intermediate bridge to facilitate high-order information propagation between multiple feature points. Hyperedge construction serves as the cornerstone of hypergraph computation. Recently, YOLOv13 defines hyperedges as learnable features, which simplifies hypergraph construction while ensuring that the model’s inference speed does not incur excessive overhead. In this work, hypergraph computation is introduced only during the training phase. Thus, leveraging the insight of adaptive hyperedge feature construction, we employ the Fuzzy K-Means algorithm~\cite{han2023vision} to perform feature clustering to get hyperegde features, \ie,
\begin{equation}
\mathcal{X}_l^{\mathcal{E}} = \text{KM}(\mathcal{X}_l) = \text{K-Means}(\mathcal{X}_l),
\end{equation}
where $\mathcal{X}_l^{\mathcal{E}} 
\in \mathbb{R}^{M_l \times C_l}$ is the hyperegde features, and $M_l$ is the number of hyperegde. To achieve multi-to-multi high-order information modeling, hypergraph convolution is conducted to perform feature aggregation and distribution. Specifically, hyperedges first collect features from all vertices and applies multi-head cross attention mechanism to refine the hyperedge feature. Then, the hyperedge features are disseminated back to the vertices to update their representations. The high-order correlations modeling can be formulated as:
\begin{equation}
\begin{aligned}
&\mathcal{R}^{hr}(\mathcal{X}_l, \mathcal{X}_l^{\mathcal{E}}) = \mathcal{R}_{\mathcal{E} \to \mathcal{V}}^{hr}(\mathcal{X}_l, \mathcal{R}_{\mathcal{V} \to \mathcal{E}}^{hr}(\mathcal{X}_l, \mathcal{X}_l^{\mathcal{E}}))  \\
\mathcal{R}_{\mathcal{V} \to \mathcal{E}}^{hr} &= \mathcal{X}_l^{\mathcal{E}}+\left[\text{SM}\left(\frac{\mathcal{X}_l^{\mathcal{E}}W_{q1}^{hr} \ (\mathcal{X}_lW_{k1}^{hr})^{\top}}{\sqrt{C_l}}\right) \mathcal{X}_lW_{v1}^{hr}\right]W_{o1}^{hr} \\
\mathcal{R}_{\mathcal{E} \to \mathcal{V}}^{hr} &= \mathcal{X}_l+\left[\text{SM}\left(\frac{\mathcal{X}_lW_{q2}^{hr} \ (\mathcal{X}_l^{\mathcal{E}}W_{k2}^{hr})^{\top}}{\sqrt{C_l}}\right) \mathcal{X}_l^{\mathcal{E}}W_{v2}^{hr}\right]W_{o2}^{hr}
\end{aligned},
\end{equation}
where $\mathcal{R}_{\mathcal{V} \to \mathcal{E}}^{hr}$ represents the massage passing from vertices to hyperedge, while where $\mathcal{R}_{\mathcal{E} \to \mathcal{V}}^{hr}$ represents the massage passing from hyperedge to vertices. $W_{q1}^{lr}\in\mathbb{R}^{C_l \times C_l}$, $W_{k1}^{lr}\in\mathbb{R}^{C_l \times C_l}$, $W_{v1}^{lr}\in\mathbb{R}^{C_l \times C_l}$ and $W_{o1}^{lr}\in\mathbb{R}^{C_l \times C_l}$ denote convolutional layers for $\mathcal{R}_{\mathcal{V} \to \mathcal{E}}^{hr}$, while $W_{q2}^{lr}\in\mathbb{R}^{C_l \times C_l}$, $W_{k2}^{lr}\in\mathbb{R}^{C_l \times C_l}$, $W_{v2}^{lr}\in\mathbb{R}^{C_l \times C_l}$ and $W_{o2}^{lr}\in\mathbb{R}^{C_l \times C_l}$ denote convolutional layers for $\mathcal{R}_{\mathcal{E} \to \mathcal{V}}^{hr}$. The loss function $\mathcal{L}_{\text{relation}}^{hr}$ of HORD is expressed as follows:
\begin{equation}
\begin{aligned}
\mathcal{L}_{\text{relation}}^{hr} &= \sum_{l \in \mathcal{L}} \sum \left(\mathcal{R}^{hr}(\mathcal{S}_l, \text{KM}(\mathcal{S}_l)) - \mathcal{R}^{hr}(\mathcal{T}_l, \text{KM}(\mathcal{T}_l))\right)^2. \\
\end{aligned}
\end{equation}
In addition, we also incorporate a multi-head attention mechanism~\cite{vaswani2017attention} to augment the diversity of the captured high-order relational correlations. It is worth noting that we do not focus on the construction process of hyper attention herein; instead, we emphasize that the concept of hypergraph computation can enhance the high-order relation transfer capability of conventional pairwise relation distillation methods. 
 
After the low-order and high-order relational distillation, the final loss function of $\text{MORD}$ in spatial domain can be expressed as:
\begin{equation}
\mathcal{L}_{\text{relation}} = \lambda_3 \cdot \mathcal{L}_{\text{relation}}^{lr} + \lambda_4 \cdot \mathcal{L}_{\text{relation}}^{hr},
\end{equation}
where $\lambda_3$ and $\lambda_4$ are weight adjustment coefficients to balance the optimization process.
\subsection{Overall loss}
\label{sec:3.4}
The overall loss function for our proposed $\text{M}^2\text{C}\text{-EvDet}$ can be formulated as:
\begin{equation}
\begin{aligned}
\mathcal{L} &= \mathcal{L}_{\text{base}} + \mathcal{L}_{\text{freq}} (\mathcal{S}, \mathcal{T}) + \mathcal{L}_{\text{relation}} (\mathcal{S}, \mathcal{T}) \\
    &= \mathcal{L}_{\text{base}} + \lambda_1 \cdot \mathcal{L}_{\text{freq}}^{lf}(\mathcal{S}, \mathcal{T}) + \lambda_2 \cdot \mathcal{L}_{\text{freq}}^{hf}(\mathcal{S}, \mathcal{T}) \\
    & \quad \ + \lambda_3 \cdot \mathcal{L}_{\text{relation}}^{lr}(\mathcal{S}, \mathcal{T}) + \lambda_4 \cdot \mathcal{L}_{\text{relation}}^{hr}(\mathcal{S}, \mathcal{T}),
\end{aligned}
\label{eq:m2c-afd}
\end{equation}
where $\mathcal{L}_{\text{base}}$ is the standard object detection loss function, which typically consists of classification loss, bounding box loss, and Intersection over Union (IoU) loss. 
\section{Experiments}
In this section, we first elaborate on the experimental settings, including datasets, implementation details, and metrics, in Sec.~\ref{sec:setting}. Subsequently, quantitative comparisons between our proposed method and state-of-the-art approaches are presented in Sec.~\ref{sec:cmp}. In Sec.~\ref{sec:abla}, ablation experiments are performed to verify the effectiveness of each proposed module and validate the generalizability of our method. Finally, several visualization results are provided in Sec.~\ref{sec:vis}.
\renewcommand{\arraystretch}{1.5} 
\begin{table*}[!htb] 
\centering
\setlength{\tabcolsep}{3mm}   
\begin{threeparttable} 
\caption{Quantitative comparison results on the DSEC-Detection dataset~\cite{gehrig2024low}. Our proposed method is compared with the event-based detection methods, and RGB-guided event-based distilled detection methods.}
    \begin{tabular}{l|ccccccc|ccc}
    \Xhline{1.1pt}   
    Detectors&Pedestrian&Rider&Car&Bus&Truck&Bicycle&Motorcycle&$\text{mAP50}$&$\text{mAP}$&Params(M) \\
    \cline{1-11}
    \multicolumn{11}{l}{\textbf{Event-based Detection}} \\
    \cline{1-11}
    $\text{DAGr}$~\cite{gehrig2024low}&-&-&-&-&-&-&-&-&14.0&- \cr
    $\text{AED}$~\cite{liu2023motion}&-&-&-&-&-&-&-&43.2&27.1&15.5 \cr
    $\text{VIT-S5}$~\cite{zubic2024state}&-&-&-&-&-&-&-&38.7&23.8&18.2 \cr
    $\text{RVT}$~\cite{gehrig2023recurrent}&-&-&-&-&-&-&-&44.2&27.7&18.5 \cr
    $\text{D-FINE-S}$~\cite{peng2024d}&15.2&37.7&42.4&20.7&17.0&32.5&7.1&42.0&24.7&10.2 \cr
    \hline
    \multicolumn{11}{l}{\textbf{RGB-guided Event-based Distilled Detection}} \\
    \cline{1-11}
    $\text{EA-DETR}$~\cite{rossi2025event}&-&-&-&-&-&-&-&27.2&14.7&- \cr
    $\text{FKD}$~\cite{zhang2020improve}&17.6&41.9&42.7&19.9&20.2&37.2&8.8&44.6&26.9&10.2 \cr
    $\text{FGD}$~\cite{yang2022focal}&17.4&43.6&42.5&23.8&20.7&36.1&7.1&45.5&27.3&10.2 \cr
    $\text{FreeKD}$~\cite{yang2022focal}&17.4&40.6&42.8&22.8&20.9&34.5&8.2&44.7&26.7&10.2 \cr
    \hline
    \rowcolor{gray!20} $\text{M}^2\text{C-EvDet (RGB)}$&39.8&58.4&61.5&45.9&36.5&53.2&33.1&69.9&46.9&10.2 \cr
    \rowcolor{gray!20} $\text{M}^2\text{C-EvDet (Event)}$&\textbf{17.8}&\textbf{44.5}&\textbf{43.0}&\textbf{24.5}&\textbf{21.8}&\textbf{37.5}&\textbf{9.0}&\textbf{45.9}&\textbf{28.3}&10.2 \cr
    \Xhline{1.1pt}
    \end{tabular}
    \label{tab1:dsec}
    \vspace{-0.3cm}
\end{threeparttable} 
\end{table*}

\subsection{Experimental Settins}
\label{sec:setting}
\textbf{Datasets.} We conduct experiments on three existing large-scale RGB-event object detection datasets, \ie, DSEC-Detection dataset~\cite{gehrig2024low}, DSEC-Det dataset~\cite{tomy2022fusing}, and PKU-DAVIS-SOD dataset~\cite{li2023sodformer}.
\begin{itemize}
    \item \textbf{DSEC-Detection} dataset~\cite{gehrig2024low} is a driving-centric dataset featuring two event cameras ($640\times 480$ pixels) and two RGB cameras ($1440\times 1080$ pixels). It provides 60 sequences (70379 frames, 390118 bounding boxes) across 8 object classes: Pedestrians, Riders, Cars, Buses, Trucks, Bicycles, and Motorcycles. 
    \item \textbf{DSEC-Det}~\cite{tomy2022fusing} dataset is another verson of DSEC dataset, and comprises 41 sequences with a total of 52,727 frames. This version emphasizes high-precision annotations for these critical classes in autonomous driving: Car, Person, and Large Vehicle. The annotations are generated on RGB frames, and then transferred to the event data through homographic transformation.
    \item \textbf{PKU-DAVIS-SOD} dataset~\cite{li2023sodformer} is a large-scale event-based detection dataset, which simultaneously records event streams and RGB frames. The dataset contains 220 sequences, comprising over 1.08 million bounding boxes, and covering three categories, \ie, Cars, Pedestrians, and Two-Wheelers.
\end{itemize}

\textbf{Implementation Details.} 
\label{implementation}
We select D-FINE~\cite{peng2024d} as the basic event-based object detection method. The proposed $\text{M}^2\text{CFD}$ module is performed in the multi-scale feature maps from PAFPN. Our model is implemented based on PyTorch and is trained on NVIDIA 4090 GPUs. We trained the model for 12 epochs with a batch size of 16 using an AdamW optimizer. The learning rate is initialized at 0.0002 and kept constant throughout the entire training process. In the $\text{AF}^2\text{D}^2$ module, we employ wavelet transforms with $T=3$ scales, and the scaling factor $r$ of AFK is set to 4. For the $\text{MORD}$ module, the spatial attenuation factor $k$ corresponding to the 3rd, 4th, and 5th-level feature maps is configured as 8, 4, and 1, respectively. We adopt the adaptive average pooling as the initialization of hyperedges and set the number of hyperedges as 16 for all levels. In terms of weight configuration, we set $\lambda_1=0.00002, \lambda_2=4, \lambda_3=0.00001, \lambda_4=0.00001$.

\textbf{Evaluation Metrics.} We choose mean average precision (\eg, COCO $\text{mAP}$~\cite{lin2014microsoft}) as the core evaluation metric, a standard benchmark extensively employed in object detection research. We give a comprehensive evaluation using average precision with various IoUs (\eg, $\text{mAP}$, $\text{mAP50}$). 

\subsection{Quantitative Comparisons}
\label{sec:cmp}
Table~\ref{tab1:dsec} presents the quantitative performance comparisons of various methods evaluated on the DSEC-Detection dataset~\cite{gehrig2024low}. Our proposed method is benchmarked against two distinct categories of baseline approaches: event-based detection methods and RGB-guided event-based distillation detection methods. The former category encompasses $\text{DAGr}$~\cite{gehrig2024low}, $\text{AED}$~\cite{liu2023motion}, $\text{VIT-S5}$~\cite{zubic2024state} and $\text{RVT}$~\cite{gehrig2023recurrent} and D-FINE~\cite{peng2024d}. The latter category comprises $\text{EA-DETR}$~\cite{rossi2025event}, $\text{FKD}$~\cite{zhang2020improve}, $\text{FGD}$~\cite{yang2022focal} and $\text{FreeKD}$~\cite{zhang2024freekd}. As can be observed from Table~\ref{tab1:dsec}, our teacher model $\text{M}^2\text{C-EvDet (RGB)}$ achieves a mAP of 46.9 and a mAP50 of 69.9, which significantly outperforms all event-based methods. This result clearly demonstrates that a substantial performance gap exists between event-only and RGB-based detection approaches. Guided by the knowledge distillation from our teacher model, our proposed method $\text{M}^2\text{C-EvDet (Event)}$ achieves state-of-the-art detection performance on the DSEC-Detection dataset. When compared with $\text{RVT}$~\cite{gehrig2023recurrent}, the leading performer among pure event-based detection methods, our approach yields a notable improvement in both mAP and mAP50, with values increasing from 27.7 and 44.2 to 28.3 and 45.9, respectively. Moreover, our model reduces the number of parameters by a significant margin of 8.3M. This result fully validates the significance and efficacy of integrating RGB semantic information into the training pipeline. From Table~\ref{tab1:dsec}, we can also observe that compared with RGB-guided event-based distilled detection methods, our method also achieves significant performance improvements on $\text{mAP}$ and $\text{mAP50}$. Specifically, compared to the second-best method, $\text{FGD}$~\cite{yang2022focal}, our method can improve the $\text{mAP50}$ and $\text{mAP}$ by 0.4 and 1, respectively. Significant performance improvements are also attained by our approach for each category. Moreover, as RGB information is introduced in the training stage for all these methods, the number of model parameters is kept constant. 

\subsection{Ablation Study}
\label{sec:abla}
\renewcommand{\arraystretch}{1.5} 
\begin{table}
\centering
\hongSwitch
\setlength{\tabcolsep}{2mm}
\begin{threeparttable} 
    \caption{Ablation experiments on the DSEC-Detection dataset~\cite{gehrig2024low}. We validate the performance of our method with and without the proposed $\text{MORD}$ and $\text{AF}^2\text{D}^2$.}
    \begin{tabular}{cc|cc|cc}
    \Xhline{1.1pt}
    \multicolumn{2}{c|}{$\text{MORD}$}&\multicolumn{2}{c|}{$\text{AF}^2\text{D}^2$}&\multirow{2}{*}{$\text{mAP50}$}&\multirow{2}{*}{$\text{mAP}$} \\
    \cline{1-4}
    Low-order&High-order&Low-freq&High-freq&& \\
    \cline{1-6}
    &&&&42.0&24.7 \cr
    \cline{1-6}
    \Checkmark&&&&43.6&26.2 \cr
    &\Checkmark&&&44.2&26.8 \cr
    \Checkmark&\Checkmark&&&44.5&27.2 \cr
    \hline
    &&\Checkmark&&44.3&26.6 \cr
    &&&\Checkmark&44.7&26.7 \cr
    &&\Checkmark&\Checkmark&44.9&27.4 \cr
    \hline
    \Checkmark&\Checkmark&\Checkmark&\Checkmark&45.9&28.3 \cr
    \Xhline{1.1pt}
    \end{tabular}
    \label{tab2:abla}
    \end{threeparttable} 
    \vspace{-0.5cm}
\end{table}

Table~\ref{tab2:abla} presents the quantitative results of the ablation experiments on the DSEC-Detection dataset~\cite{gehrig2024low}. We utilize variants of the model with and without the low-order, high-order, low-frequency and high-frequency components to verify the effectiveness of $\text{MORD}$ and $\text{AF}^2\text{D}^2$. 

\textbf{Low-Order and High-order Relational Distillation.} It can be seen from Table~\ref{tab2:abla} that adding low-order and high-order relational distillation separately both contributes to the enhancement of mAP. Specifically, low-order relational distillation increases the mAP from 24.7 to 26.2, while high-order relational distillation elevates the mAP to 26.8. These comprehensive performance improvements verify the significance of low-order and high-order relationships in relational distillation. Furthermore, when both low-order and high-order relational distillation are adopted simultaneously, the mAP and mAP50 can be further improved to 27.2 and 44.5, respectively, which are higher than those achieved by either module alone. This phenomenon fully demonstrates that low-order and high-order relationships are not conflicting, but rather two complementary distillation paradigms, which jointly model the abundant and complex object relationships in multi-modal images and effectively accomplish cross-modal knowledge transfer. 

\hong{\textbf{Low-frequency and High-frequency Feature Distillation.} It can be seen from Table~\ref{tab2:abla} that adding low-frequency and high-frequency feature distillation separately both contribute to the enhancement of mAP. Specifically, low-frequency distillation increases the mAP from 24.7 to 26.6, while high-frequency distillation elevates the mAP to 26.7. These comprehensive performance improvements verify the significance of low-frequency and high-frequency distillation. Furthermore, when both low-frequency and high-frequency feature distillation are adopted simultaneously, the mAP and mAP50 can be further improved to 27.4 and 44.9, respectively, which are higher than those achieved by either module alone. To further verify the effectiveness of frequency-domain decoupling, we additionally conduct comparison experiments on distillation with and without frequency decoupling, as shown in Table~\ref{tab3:whole}. Specifically, the PSD refers to the plain spatial-domain distillation without frequency decoupling. It can be observed that the mAP of PSD is only 25.8, which is far lower than that of high- and low-frequency decoupled distillation. This phenomenon fully demonstrates that our $\text{AF}^2\text{D}^2$ can alleviate the aliasing between modality-agnostic and modality-specific knowledge.}
\renewcommand{\arraystretch}{1.5} 
\begin{table}
\centering
\hongSwitch
\setlength{\tabcolsep}{4mm}   
\begin{threeparttable} 
    \caption{Comparison results between the plain spatial distillation without frequency decoupling and the proposed $\text{AF}^2\text{D}^2$ module.}
    \begin{tabular}{cc|cc}
    \Xhline{1.1pt}
    Method&Frequency Decoupling&mAP50&mAP \cr
    \cline{1-4}
    PSD&&43.5&25.8 \cr
    $\text{AF}^2\text{D}^2$&\Checkmark&44.9&27.4 \cr
    \Xhline{1.1pt}
    \end{tabular}
    \label{tab3:whole}
    \end{threeparttable} 
    \vspace{-0.5cm}
\end{table}

\textbf{Multi-Domain Multi-order Cross-modal Distillation} Table~\ref{tab2:abla} show that the integration of $\text{MORD}$ and $\text{AF}^2\text{D}^2$ leads to continuous improvement, with the final results reaching 28.3 for mAP and 45.9 for mAP50. This phenomenon arises because $\text{MORD}$ and $\text{AF}^2\text{D}^2$ are knowledge transfer methods from two distinct perspectives: the former focuses on the object semantic information while the latter emphasizes the complex correlation relationships among objects.

\begin{figure}
    \centering  
    \subfloat[\footnotesize $\lambda_1$ and $\lambda_2$]{
        \includegraphics[width=0.5\linewidth]{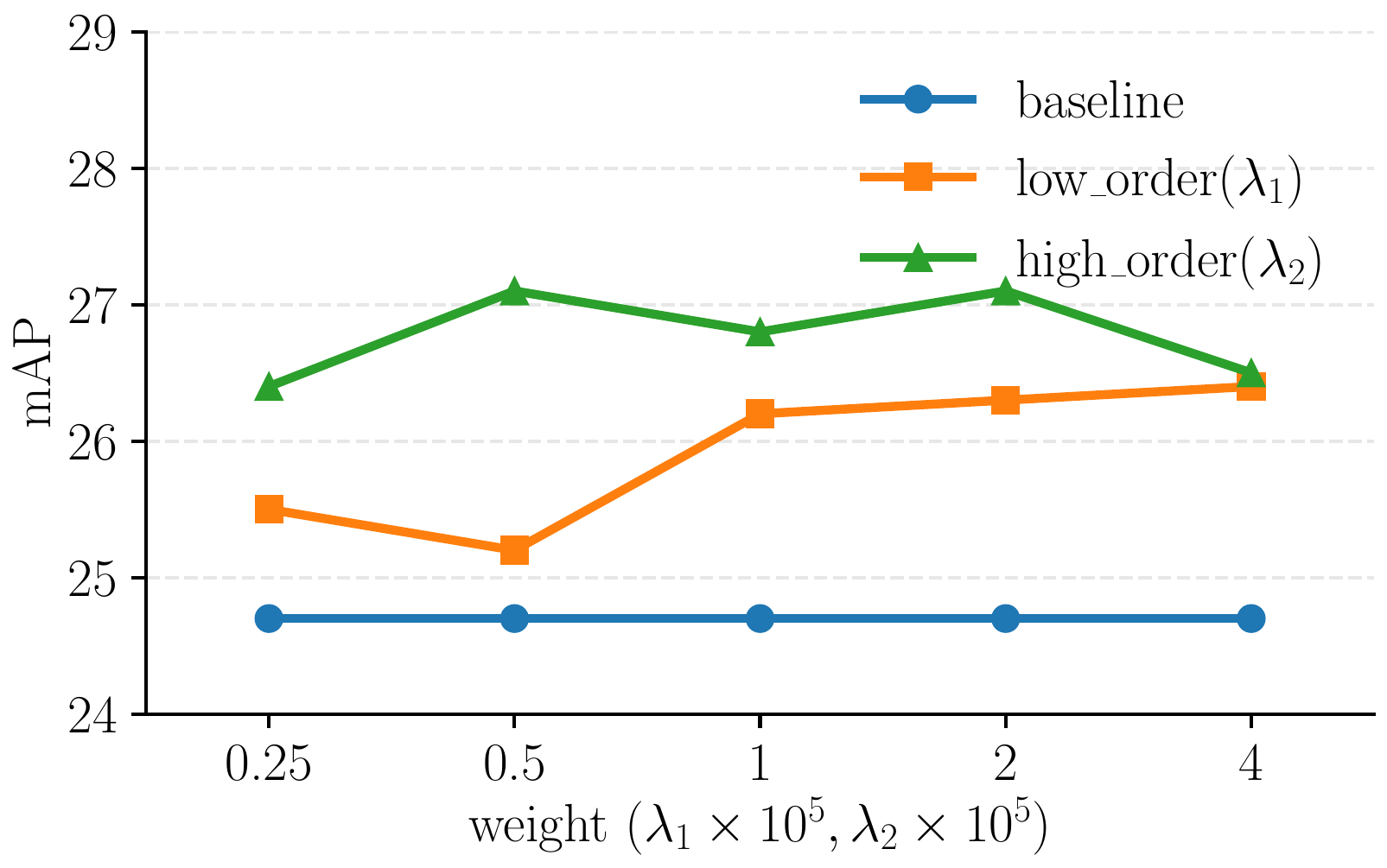}%
        \label{fig5:param1}
    }
    \hfill  
    \subfloat[\footnotesize $\lambda_3$ and $\lambda_4$]{%
        \includegraphics[width=0.5\linewidth]{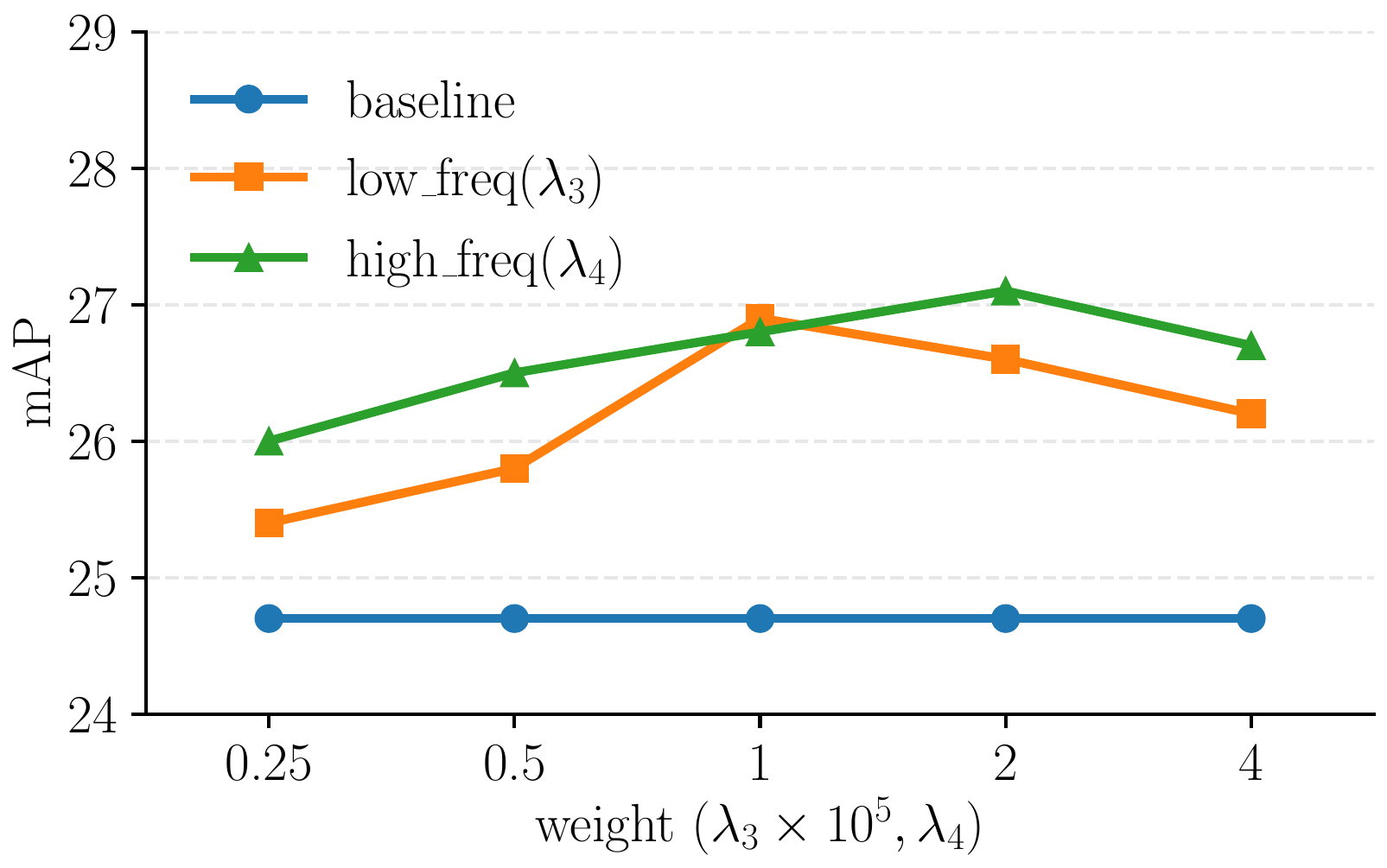}%
        \label{fig5:param2}
    }
    \caption{Experimental results of sensitivity analysis of distillation loss weight on the DSEC-Detection dataset~\cite{gehrig2024low}.}
    \label{fig5:param}  
\end{figure}

\textbf{Parameter Sensitivity Analysis.} Fig.~\ref{fig5:param} shows the performance of our proposed method under different distillation loss weights. From the Fig.~\ref{fig5:param1}, we can observe that when the $\text{MORD}$ loss weights $\lambda_1$ and $\lambda_2$ changes from $0.25 \times 10^{-5}$ to $4 \times 10^{-5}$, the $\text{mAP}$ remains basically stable, ranging from 25.2 to 27.1. In addition, we find that the effectiveness of high-order relational distillation is superior to that of low-order relational distillation, which also indicates that multi-to-multi relationships are more widely prevalent in visual object detection tasks, and our hyper-attention can effectively capture these relationships. As for the $\text{AF}^2\text{D}^2$ module in Fig.~\ref{fig5:param2}, when $\lambda_3$ changes from $0.25 \times 10^{-5}$ to $4 \times 10^{-5}$ and $\lambda_4$ changes from $0.25$ to $4$, the model's performance remains stable, oscillating between 25.4 and 27.1, which indicates that our method is insensitive to its weight. 

\hong{\textbf{The Initialization and the Number of Hyperedges.} We conduct extensive experiments to verify the sensitivity of the MORD module to initialization schemes and the number of hyperedges. For hyperedge initialization, we evaluate four strategies: adaptive average pooling (AdaAvg.), adaptive max pooling (AdaMax.), and random initialization with different seeds (Rand. 1 and 2). As shown in Table~\ref{tab4:hyperedge}, our method is highly insensitive to the initialization of the K-Means clustering algorithm, with mAP values maintained in a narrow range from 28.0 to 28.3, which validates the stability of our MORD module. Regarding the number of hyperedges $M$, we test $M = 4, 9, 16, 25$. The results show that performance gradually improves with increasing $M$ and achieves the peak mAP of 28.3 at $M = 16$. Further increasing $M$ leads to performance saturation. Accordingly, we set the number of hyperedges to 16 to strike a balance between computational efficiency and relational modeling capacity.}
\renewcommand{\arraystretch}{1.5} 
\begin{table}
\hongSwitch
\vspace{-0.3cm}
\label{tab:dsec}
	\centering
	\setlength{\tabcolsep}{3mm}   
	\begin{threeparttable} 
		\caption{Experimental results of sensitivity analysis of initialization and number of hyperedge on the DSEC-Detection dataset~\cite{gehrig2024low}.}
		\begin{tabular}{ccc|ccc}
		\Xhline{1.1pt}  
        \multicolumn{3}{c|}{Hyperedge Initialization}&\multicolumn{3}{c}{Hyperedge Number} \\
        \cline{1-6}
        Mode&mAP50&mAP&$M$&mAP50&mAP \\
        \cline{1-6}
        AdaAvg.&45.9&28.3&4&45.3&27.8 \cr
        AdaMax.&46.0&28.1&9&45.7&28.0 \cr
        Rand. 1&45.6&28.1&16&45.9&28.3 \cr
        Rand. 2&45.8&28.2&25&45.7&28.2 \cr
	\Xhline{1.1pt}
	\end{tabular}
    \label{tab4:hyperedge}
	\end{threeparttable} 
\end{table}

\hong{\textbf{Robustness Analysis for $\text{AF}^2\text{D}^2$ module.} In real-world scenarios, RGB-Event datasets are prone to modal noise and spatial misalignment, which pose significant challenges to the robustness of detection methods. To empirically demonstrate this robustness, we conducted sensitivity experiments on the DSEC-Detection dataset~\cite{tomy2022fusing} by introducing random spatial offsets and random Gaussian noise in the training process. We let $p$ denote the range of spatial offsets and $r$ represent the kernel radius of Gaussian noise. The results presented in Table~\ref{tab5:robust} demonstrate that our method preserves superior performance under small-scale perturbations. For instance, with an 8-pixel random spatial offset, our model still attains 44.8 mAP50 and 27.2 mAP, respectively. When interference becomes extreme (i.e., $p$ and $r$ become larger), the performance starts to decline. This phenomenon is reasonable and fully validate the robustness and practical stability of the proposed $\text{AF}^2\text{D}^2$.}
\renewcommand{\arraystretch}{1.5} 
\begin{table}
\hongSwitch
\vspace{-0.3cm}
\label{tab:dsec}
	\centering
	\setlength{\tabcolsep}{4mm}   
	\begin{threeparttable} 
		\caption{Robustness experiment for $\text{AF}^2\text{D}^2$ module on DSEC-Detection dataset~\cite{gehrig2024low}.}
		\begin{tabular}{ccc|ccc}
		\Xhline{1.1pt}  
        \multicolumn{3}{c|}{Random Spatial Shift}&\multicolumn{3}{c}{Random Gaussian Noise} \\
        \cline{1-6}
        $p$&mAP50&mAP&$r$&mAP50&mAP \\
        \cline{1-6}
        0&44.9&27.4&0&44.9&27.4 \cr
        4&45.0&27.3&2&45.0&27.4 \cr
        8&44.8&27.2&3&44.7&27.2 \cr
        16&44.5&26.8&5&43.9&26.6 \cr
	\Xhline{1.1pt}
	\end{tabular}
    \label{tab5:robust}
	\end{threeparttable} 
\end{table}

\hong{\textbf{Performance on Challenging Scenarios.} To further verify the effectiveness of our methods in challenging scenarios, we selected 1220 low-quality frames from the DSEC-Detection~\cite{tomy2022fusing} to construct a subset, named Challenging-DSEC-Detection. This set includes 420 frames with severe motion blur (the $interlaken\_00\_b$ folder) and 800 frames from low-light night sequences (the $zurich\_city\_12\_a$ folder). As shown in the performance comparison in Table~\ref{tab6:challenging}, $\text{AF}^2\text{D}^2$ still achieves an improvement of 1.9 mAP50 and 1.3 mAP on the Challenging-DSEC compared to the baseline. These results fully verify that $\text{M}^2\text{C-EvDet}$ possesses the ability to enhance detection performance under motion blur or low-light scenarios, demonstrating strong adaptability to degraded guidance. We also observe that the improvement margin is relatively lower compared to the entire DSEC-Detection dataset, which inspires us to further expand high-quality RGB-guided event datasets and design more advanced methods for transferring knowledge from low-quality teacher images.}
\renewcommand{\arraystretch}{1.5} 
\begin{table}
\hongSwitch
\vspace{-0.3cm}
\label{tab:dsec}
	\centering
	\setlength{\tabcolsep}{2mm}   
	\begin{threeparttable} 
		\caption{Comparison results on challenging and overall scenarios of DSEC-Detection dataset~\cite{gehrig2024low}.}
		\begin{tabular}{c|cc|cc}
		\Xhline{1.1pt}  
        \multirow{2}{*}{Detectors}&\multicolumn{2}{c|}{Challenging-DSEC-Detection}&\multicolumn{2}{c}{DSEC-Detection} \\
        \cline{2-5}
        &mAP50&mAP&mAP50&mAP \\
        \cline{1-5}
        Baseline&25.5&13.1&42.0&24.7 \cr
        $\text{M}^2\text{C-EvDet}$&27.4&14.4&45.9&28.3 \cr
	\Xhline{1.1pt}
	\end{tabular}
    \label{tab6:challenging}
	\end{threeparttable} 
\end{table}

\textbf{Performance on Other Datasets.} To validate the generalization capability of our proposed approach, we also conduct experiments on two additional datasets, namely DSEC-Det dataset~\cite{tomy2022fusing} and PKU-DAVIS-SOD~\cite{li2023sodformer}. The detection results on these two datasets are presented in the Table~\ref{tab7:gen}, where ‘T.W.’ and ‘L.V.’ refer to ‘Two-Wheels’ and ‘Large-Vehicle’, respectively. It can be observed that our method still achieves remarkable performance improvements, with the mAP increased by 0.7 and 2.8, respectively. Furthermore, the performance of each category has also been significantly improved, which further verifies the effectiveness of our method in more realistic scenarios.
\renewcommand{\arraystretch}{1.5} 
\begin{table}[!b] 
    \vspace{-0.3cm}
\label{tab:dsec}
	\centering
	\setlength{\tabcolsep}{0.8mm}   
	\begin{threeparttable} 
		\caption{Detection performance on DSEC-Det~\cite{tomy2022fusing} dataset and PKU-DAVIS-SOD~\cite{li2023sodformer} dataset.}
		\begin{tabular}{c|ccc|c|ccc|c}
		\Xhline{1.1pt}  
        \multirow{2}{*}{Detectors}&\multicolumn{4}{c|}{PKU-DAVIS-SOD}&\multicolumn{4}{c}{DSEC-Det} \\
        \cline{2-9}
        &Car&Pedestrian&T.W.&mAP&Car&Person&L.V.&mAP \\
        \cline{1-9}
        Baseline&31.5&10.4&26.4&22.7&41.7&12.9&30.2&28.3 \cr
        $\text{M}^2\text{C-EvDet}$&31.9&11.4&27.4&23.5&43.6&15.7&34.0&31.1 \cr
	\Xhline{1.1pt}
		\end{tabular}
    \label{tab7:gen}
	\end{threeparttable} 
\end{table}

\hong{\textbf{Performance on Other Detection Architectures.} To further demonstrate the broad generalization capability of our distillation strategy across various architectures, we conducted additional experiments using the YOLOv8~\cite{Jocher_Ultralytics_YOLO_2023} and GFL~\cite{li2020generalized} frameworks. As presented in Table~\ref{tab8:yolo}, the results consistently show performance gains across different backbones. With the proposed distillation strategy, YOLOv8 achieves an improvement of 2.7 and 2.8 in mAP50 and mAP, respectively, while GFL obtains corresponding gains of 3.1 and 2.8. All these performance improvements fully validate the generalization ability of our proposed method.}
\renewcommand{\arraystretch}{1.5} 
\begin{table}
\hongSwitch
\vspace{-0.3cm}
\label{tab:dsec}
	\centering
	\setlength{\tabcolsep}{3mm}   
	\begin{threeparttable} 
		\caption{Generalization experiment with different detection architectures and backbone network on DSEC-Detection dataset~\cite{gehrig2024low}.}
		\begin{tabular}{ccc|cc}
		\Xhline{1.1pt}  
        Detectors&Backbone&Distillation&mAP50&mAP \\
        \cline{1-5}
        \multirow{2}{*}{YOLOv8s}&\multirow{2}{*}{Darknet53}
        &&39.9&22.8 \\
        &&\Checkmark&42.6&25.6 \cr
        \cline{1-5}
        \multirow{2}{*}{GFL}&\multirow{2}{*}{ResNet50}
        &&60.1&38.2 \\
        &&\Checkmark&63.2&41.0 \cr
	\Xhline{1.1pt}
	\end{tabular}
    \label{tab8:yolo}
	\end{threeparttable} 
\end{table}

\subsection{Visualization Results}
\label{sec:vis}
\begin{figure}
\begin{center}
\includegraphics[width=1.0\linewidth]{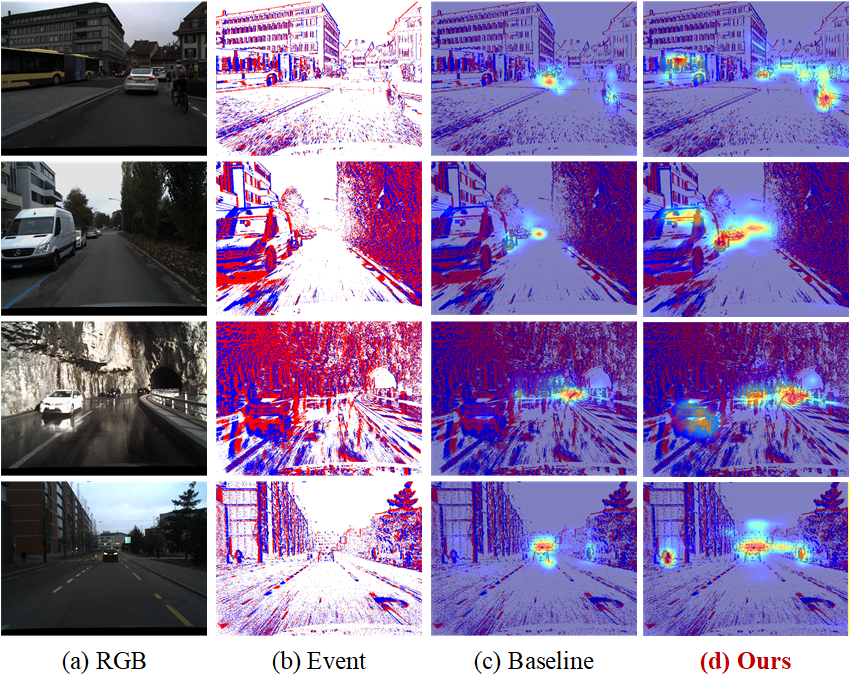}
\end{center}
\vspace{-0.3cm}
\caption{Visual comparison of feature activation maps between the baseline and $\text{M}^2\text{C-EvDet}$ on the DSEC-Detection dataset~\cite{gehrig2024low}.}
\label{fig6:feat}
\vspace{-0.3cm}
\end{figure}

\textbf{Visualization of Feature Maps}. We adopt the Grad-CAM visualization method to intuitively present the feature activation maps of our method across diverse scenarios. Fig.~\ref{fig6:feat} displays the RGB image, event representation, features of the baseline, and features of our $\text{M}^2\text{C-EvDet}$ from left to right, respectively. It can be observed that the features of the baseline merely cover partial regions in a coarse manner and fail to focus on target areas, resulting in suboptimal performance under complex scenarios such as low-light conditions and high-speed motion. In contrast, the features of our method exhibit remarkable advantages, as they can accurately locate the latent semantic regions that contain the objects of interest, validating the effectiveness of transferring frequency-domain semantics and multi-order correlations from RGB images.

\begin{figure}
\begin{center}
\includegraphics[width=1.0\linewidth]{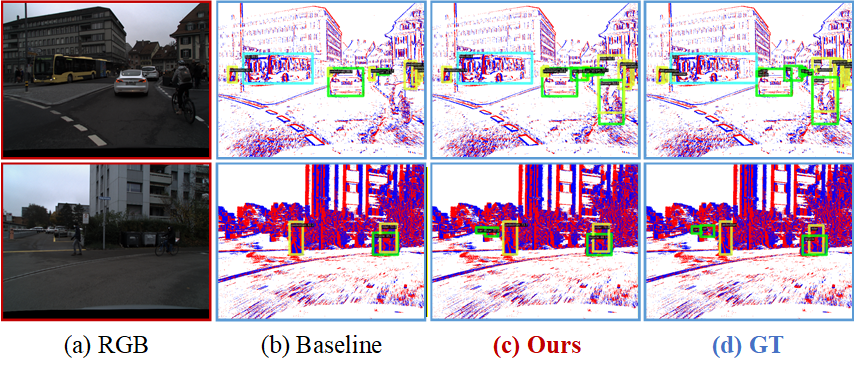}
\end{center}
\vspace{-0.3cm}
\caption{Visual comparison of detection results between the baseline and $\text{M}^2\text{C-Det}$ on the DSEC-Detection dataset~\cite{gehrig2024low}.}
\label{fig7:vis}
\vspace{-0.3cm}
\end{figure}

\textbf{Visualization of Detection Results}. 
Fig.~\ref{fig7:vis} presents the RGB image, detection results of the baseline method, outputs of our proposed $\text{M}^2\text{C-EvDet}$, and the Ground Truth (GT) annotations from left to right, respectively. It is evident from the visualization that the baseline method—plagued by insufficient texture details and inadequate semantic feature learning—exhibits substantial missed detections of objects in traffic intersection scenarios, thereby restricting its real-world applicability. On the contrary, our approach integrates object semantic representations and high-order object correlations in the training pipeline, which allows the model to exploit contextual scene information for improved detection accuracy and holds significant promise for practical deployment in safety-critical scenarios such as autonomous driving.

\begin{figure}
\begin{center}
\includegraphics[width=1.0\linewidth]{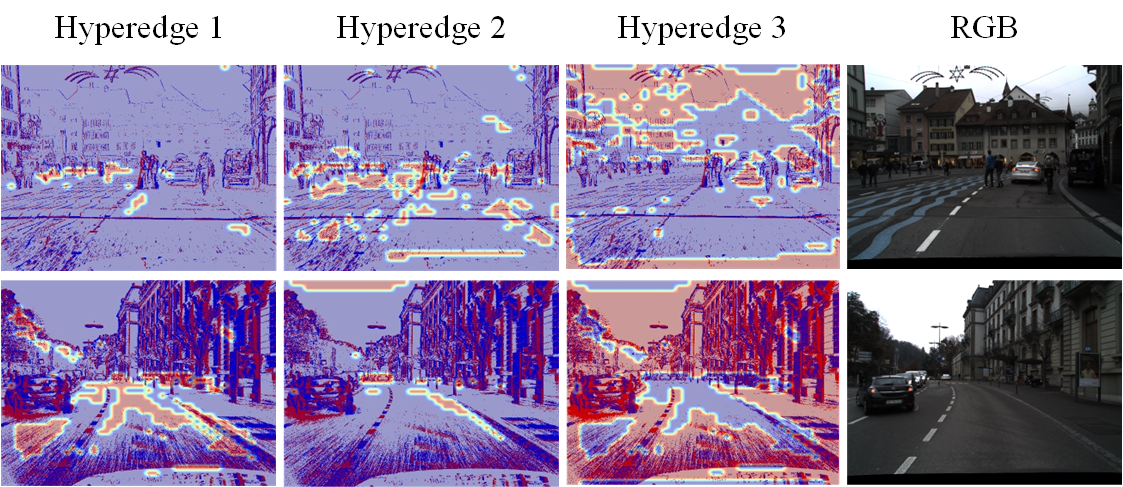}
\end{center}
\vspace{-0.3cm}
\caption{\hong{Visualization of adaptive hyperedges for MORD module on the DSEC-Detection dataset~\cite{gehrig2024low}.}}
\label{fig8:hyperedge}
\vspace{-0.3cm}
\end{figure}

\hong{\textbf{Visualization of Adaptive Hyperedges}. Fig.~\ref{fig8:hyperedge} illustrates the distribution of hyperedges in scenarios characterized by an extensive density of small objects. It can be observed that the proposed MORD method constructs hyperedge connections across various semantic correlations, enabling precise discrimination among "person," "motorcycle," and "car" categories in crowded scenes via high-order information propagation. Furthermore, comparing the hyperedge distributions in the first and second columns of the first row in Fig.~\ref{fig8:hyperedge} reveals that MORD facilitates finer-grained high-order interactions between distinct instances of the same category. This further strengthens the high-order knowledge distillation capability and enhances detection performance in dense small-object scenarios.}

\begin{figure}
\begin{center}
\includegraphics[width=1.0\linewidth]{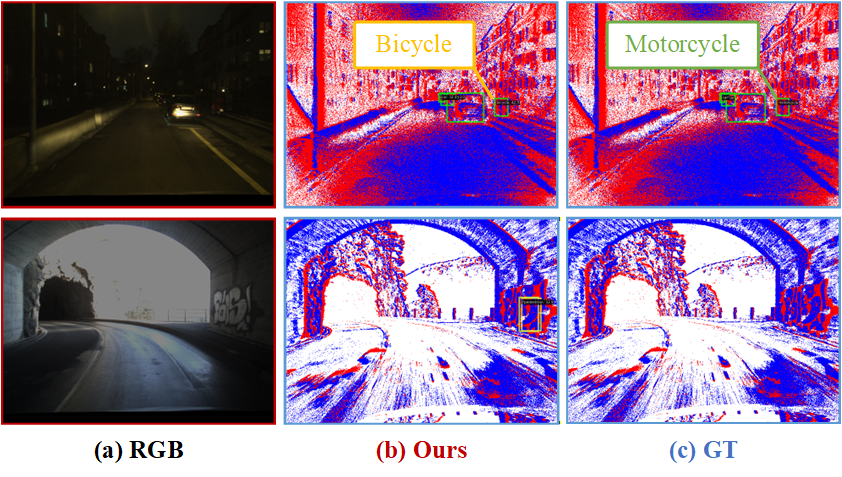}
\end{center}
\vspace{-0.3cm}
\caption{\hong{Visual results of failure case for $\text{M}^2\text{C-Det}$ on the DSEC-Detection dataset~\cite{gehrig2024low}.}}
\label{fig9:badcase}
\vspace{-0.3cm}
\end{figure}

\hong{\subsection{Limitation and Failure Case Analysis}
$\text{M}^2\text{C-EvDet}$ leverages the rich visual and texture information inherent in the RGB modality to fundamentally enhance the feature extraction capability of the event data. Our methods mitigates the cross-modal gap but cannot completely eliminate modal discrepancies, due to the low-quality RGB images and the inadequate model optimization process. As illustrated in the low-light scenarios of Fig.~\ref{fig9:badcase}, while our relational distillation successfully enables the model to detect objects, the lack of sufficient texture information in the RGB teacher can lead to semantic confusion. For example, motorcycles are occasionally misclassified as bicycles due to the extreme loss of discriminative features. Similarly, in overexposed scenarios, our method generates several false positives due to inadequate information transfer. These failure cases also point out the direction for us to design more effective transfer methods and improve the generalizability and practicality of the proposed framework in future work.}
\section{Conclusion}
This paper introduces $\text{M}^2\text{C-EvDet}$, a RGB-guided EvOD method that pioneers the limitations of existing cross-modal knowledge distillation methods that only focus on single spatial semantics or pairwise relations via frequency learning and hypergraph computation. Specifically, we propose $\text{AF}^2\text{D}^2$ , which breaks the limitations of conventional spatial-domain distillation by adaptively decoupling low-frequency and high-frequency components. This design effectively alleviates the aliasing between modality-agnostic and modality-specific knowledge, while boosting the transfer of both global and local object semantics. $\text{MORD}$ extends the scope of relational distillation by modeling both pairwise low-order relations and multi-to-multi high-order relational structures. To capture high-order relations, we design a hyper-attention mechanism that employs K-means clustering for hyperedge feature construction, facilitating the aggregation-and-distribution paradigm among multiple feature points. Extensive experiments on three datasets validate that $\text{M}^2\text{C-EvDet}$ achieves significant performance improvements and establishes a new state-of-the-art benchmark. Future research will focus on optimizing the efficiency of hypergraph construction and developing logit distillation techniques specifically tailored for cross-modal information transfer, with the goal of further improving the generalizability and practicality of the proposed framework.
\bibliographystyle{IEEEtran}
\bibliography{ref}
\end{document}